\documentclass[twoside,11pt]{article}

\usepackage{jmlr2e}
\usepackage{graphicx}
\usepackage{color}
\usepackage{amsmath}
\usepackage{amssymb}
\usepackage{hyperref}

\ShortHeadings{MACEst: The reliable and trustworthy Model Agnostic Confidence Estimator}{Green, Rowe and Polleri}
\firstpageno{1}

\begin{document}

\title{MACEst: The reliable and trustworthy Model Agnostic Confidence Estimator}

\author{\name Rhys Green \email greenr10@cardiff.ac.uk \\
       \addr Gravity Exploration Institute\\
       Cardiff University \\
       Cardiff, UK
       \AND
       \name Matthew Rowe \email matthew.r.rowe@oracle.com \\
       \addr Oracle AI Apps \\
       Oracle \\
       London, UK
       \AND
       \name Alberto Polleri \email alberto.polleri@oracle.com\\
       \addr Oracle AI Apps\\
       Oracle \\
       London, UK}

\editor{ }

\maketitle

\begin{abstract}
 Reliable Confidence Estimates are hugely important for any machine learning model to be truly useful. 
 In this paper we argue that any confidence estimates based upon standard machine learning point prediction algorithms are fundamentally flawed and under situations with a large amount of epistemic uncertainty are likely to be untrustworthy. 
 To address these issues, we present MACEst, a Model Agnostic Confidence Estimator, which provides reliable and trustworthy confidence estimates. 
 The algorithm differs from current methods by estimating confidence \emph{independently} as a local quantity which explicitly accounts for both aleatoric and epistemic uncertainty. 
 This approach differs from standard calibration methods that use a global point prediction model as a starting point for the confidence estimate.
\end{abstract}


\section{Introduction}
Over recent decades a huge amount of progress have been made on improving the global accuracy of machine learning models, however calculating how \emph{likely} a single prediction is to be correct has seen considerably less attention. 
In some fields, where making a single bad prediction can have major consequences, having trustworthy confidence estimates may be the limiting factor before introducing AI. 
It is important in these situations that a model is able to understand how likely any prediction is to be correct before acting upon it; being able to do this well requires satisfying two closely related conditions:

\begin{enumerate}
    \item Confidence estimates must approximate the true relative frequency of being correct, i.e. if a model estimates a confidence of 80\% it should be correct roughly 80\% of the time.
    \item Confidence estimates should also indicate \emph{ignorance}, i.e. the model must know what it doesn't know so that it will not blindly make bad predictions.
\end{enumerate}

A similar way to think about confidence estimation is to say that any estimate must account for any \emph{uncertainty} that is present. 
Uncertainty can be split into two forms (\cite{lakshminarayanan2017simple, kendall2017uncertainties}):

\begin{enumerate}
    \item Aleatoric Uncertainty: this refers to the intrinsic variance or randomness inherent in any process, i.e. even with unlimited data there will always be errors in any modelling and therefore aleatoric uncertainty cannot be reduced by collecting more data.
    \item Epistemic Uncertainty: this refers to the uncertainty due to the lack of knowledge of a model, the knowledge of any model comes from data; i.e. does the model have the \emph{relevant} data to predict something? 
    Epistemic Uncertainty can sometimes be reduced by collecting more or different data. 
\end{enumerate}

\begin{figure}[t]
    \includegraphics[width=0.95 \textwidth]{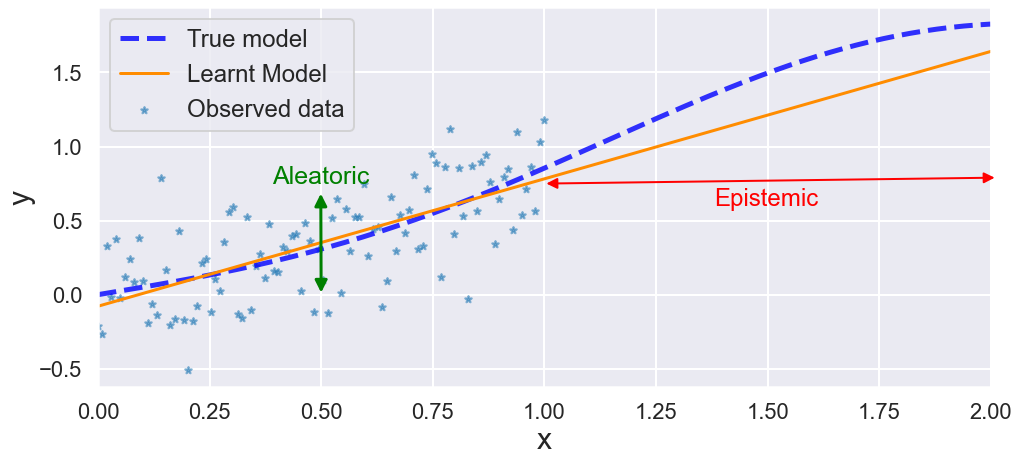}
    \centering
    \caption{A simple example illustrating the two types of uncertainty: Aleatoric refers to the uncertainty due to the intrinsic randomness inherent in any system. 
    This is shown by the spread of observed data about the learnt model which is a good approximation to the true model in the region we have observed data. 
    Epistemic uncertainty is the uncertainty due to a lack of knowledge (i.e data) to inform the model's predictions. 
    This is shown above when the model becomes a worse approximation to the true function as we move further away from the observed data. (e.g $x > 1)$ }
\end{figure}

Understanding and accounting for uncertainty is crucial to understanding the capabilities and limitations of any machine learning model, but it is especially crucial if one wants to produce a confidence estimate.
Confidence estimation for a prediction is therefore most naturally expressed in a Bayesian language where uncertainty can be explicitly accounted for in both the data and the model. 
Once we have accounted for this uncertainty we can then reason about the effect it has upon a prediction; this is in general done by marginalising out our uncertainty to produce a distribution of outputs for any prediction that will then reflect our confidence (\cite{BishopPatternrecognitionmachine2006}).

Despite the Bayesian framework being the natural way to tackle the problem of confidence estimation, it is often prohibitively computationally expensive in its application. 
It also does not naturally lend itself to being combined with popular machine learning algorithms which have a large number of parameters or non-linearities such as Decision trees, Random Forest, support vector machine etc. (\cite{loh2014fifty, breiman2001random, cortes1995support})

Most popular machine learning algorithms adopt methods to produce an estimate of how likely a prediction is to be correct, for example a random forest model (\cite{breiman2001random}) estimates the confidence score by the fraction of trees that predict a certain class. 
Each of such techniques are in effect \emph{heuristics} which correspond to ordering predictions that are more or less likely, however they certainly cannot be interpreted as true probability or confidence estimates.

Although these heuristics can be useful, and when calibrated can estimate the aleatoric uncertainty well (see \ref{sec:Calibration}), we argue that they will always be fundamentally flawed when estimating confidence due to using the classification model estimate as a starting point - see Section \ref{sec: not trustworthy} for a more detailed analysis of this claim.

In \cite{JiangTrustNotTrust2018} the authors present a method which provides Trust scores that attempt to quantify how much one can trust a classifier, this is very closely related to the notion of our confidence in being correct. 
These scores are computed explicitly by calculating the relative distance to a set of $k$ nearest neighbours from each class, which can be seen as quantifying the epistemic uncertainty for a given prediction.

In this paper we highlight the fundamental problem with using point prediction algorithms (even if well calibrated to data) and present an alternative model MACEst (Model Agnostic Confidence Estimator) that seeks to bridge the gap between trust scores and the current state of the art for confidence calibration. 
We show that for many tasks our method is competitive with the state of the art and, as it can directly account for both aleatoric and epistemic uncertainty, is more robust to problems such as extrapolation and out of sample bias.  
\newline
\newline
The original contributions of this work are as follows: 

\begin{itemize}
    \item A comparison of trust scores and confidence calibration highlighting key similarities and differences between these methods.
    \item Demonstration of how calibration methods in general do not properly account for epistemic uncertainty and therefore can return high confidence but a low trust score.
    \item A novel algorithm called MACEst that is agnostic to the model which produces point predictions and which explicitly models both aleatoric and epistemic uncertainty. 
    \item Experimental evidence of MACEst's competitive calibration performance with respect to popular calibration, measured across a range of metrics; and its robustness to large epistemic uncertainty, where existing calibration methods perform poorly.
\end{itemize}

An implementation of the method as well as code to reproduce the results of the paper is freely available at at https://github.com/oracle/macest.

\section{Confidence Estimation }

The general approach of classification involves using labelled pairs of training data $(\mathbf{X},\mathbf{y}) $ where each $\mathbf{x_i} \in \mathbf{X}$ is a vector corresponding to features, and each $\mathbf{y_i} \in \mathbf{Y}$ is an associated class for the element $i$: the goal being to learn a transformation function $\psi(\mathbf{X}) $ that maps any new unseen data point $\mathbf{x_*}$ to an estimated label $\hat{y_*}$. 

The problem can then be interpreted as wanting to estimate the class label $y_*$ of $ \mathbf{x_*} $ given a learnt model $\psi(\mathbf{X})$ from training data $\mathbf{X}$. 
It is important to highlight that, although somewhat obvious, any prediction explicitly depends upon the specific model which has been trained using a specific set of training data $\mathbf{X}$. 
For a given point $\mathbf{x_*} $ we estimate the confidence as:

\begin{equation}\label{eq:full_prob}
    C(\hat{y_*} = y_* |\mathbf{x_*}, \psi(\mathbf{x_*}), \mathbf{X}) = \hat{p_*}
\end{equation}

Where $y_*$ is the true class label, $\hat{y_*}$ is the predicted class label for a point $ \mathbf{x_*} $ and $\hat{p_*}$ is the estimated confidence associated with that prediction. 
This then corresponds to estimating the probability that our prediction $\hat{y}$ is correct given a model ($\psi$) learnt from the training data $(\mathbf{X},\mathbf{y})$. 

Note that in this work we will be looking at \emph{Confidence estimation}, therefore we only look at the confidence of a point prediction being correct rather than estimating the full distribution across all classes (these are equivalent for binary classification but not multi-class) - this is similar to the definitions in \cite{GuoCalibrationModernNeural2017}. 
We argue that this is the more natural question to be answered by a confidence estimator given an already trained point prediction model, it is also more similar to the notion of Trust Scores defined in \cite{JiangTrustNotTrust2018} to which we would like to draw parallels. 

The confidence estimator $C$ defined above must also explicitly account for the training data and, more importantly, how \emph{informative} the training data is when trying to classify a specific point, $x_*$. 
This notion is the foundation of Gaussian Process modelling (\cite{rasmussen2003gaussian, williams2006gaussian}) where the confidence of any prediction is calculated by explicitly considering the kernel (or co-variance) function $K(\mathbf{x_*}, \mathbf{X})$, where the kernel can be interpreted as a measure of similarity between a point $ \mathbf{x_*}$ and the training data $\mathbf{X}$. 
Here, the confidence of any prediction is conditioned upon the similarity between the predicted point and the training data: if a point is \emph{not} similar to what has been seen during training then we \emph{cannot} be confident when predicting it. 

Having set out what we mean by a confidence estimate (equation \ref{eq:full_prob}), for brevity and ease of comparison of notation with other literature, we will also henceforth adopt the convention of, whilst of course not ignoring, leaving the dependence upon training data and model implicit. 
We therefore define the confidence of a given point prediction being correct as: 

\begin{equation}\label{eq:small_prob}
    C(\hat{y_*} = y_* ) = \hat{p_*}
\end{equation} 
where again we define $p$ as the confidence of being correct, $y_*$ as the true label, and $\hat{y_*}i$ as the predicted label.

\section{Performance Metrics}
The calibration of a confidence estimator is the degree to which the estimates match the empirical accuracy of a classifier, i.e. if a confidence estimator estimates a probability of 75\%, this should be correct approximately 75\% of the time. 
We say that it is \emph{perfectly calibrated} if this is true for all estimated probabilities, $p \in [0,1]$.
If this is not the case then we say that the confidence estimator is either \emph{over} or \emph{under} confident: i.e. if it is correct less often than the estimated probability an estimator is said to be over-confident and vice versa.

Formally, this can be defined as needing to satisfy the following condition:

\begin{equation}
    P(\hat{y_i} = y_i | \hat{p_i} = p_i) = p   \quad  \forall p [0,1]
\end{equation}

As the true probability of an event is an unknown random variable, it is generally not possible to calculate this exactly, so the standard technique is to convert the continuous confidence space into a set of $n$ discrete bins, $B_n$ (e.g $[0.5,0.6, 0.7,0.8, 0.9, 1.]$) and compare the average confidence estimate $\bar{p}$ of each bin to the empirical accuracy of the bin. 

That is:
\begin{equation}
    Acc(B_n) = \frac{1}{|B_n|} \sum_{i\in B_n} \large{\delta_{\hat{y_i}, y_i}}
\end{equation}

Where,
\begin{equation}
  \delta_{\hat{y}, y} =\left\{
  \begin{array}{@{}ll@{}}
    1, & \text{if}\ \hat{y} =y \\
    0, & \text{otherwise}
  \end{array}\right.
\end{equation} 

And,
\begin{equation}
    C(B_n) = \frac{1}{|B_n|} \sum_{i\in B_n} \bar{p_i}
\end{equation}

A perfectly calibrated estimator will then have $C(B_n) = Acc(B_n) $ for each bin. 



The standard metric when looking to evaluate the calibration error directly is Expected Calibration Error (ECE) \citep{naeini2015obtaining}. 
This metric measures the difference between the predicted confidence estimates and the empirical accuracy of each bin. 
These residuals are then combined in a sum weighted by the number of points in each bin:
\begin{equation} \label{eq: calibration_equation}
    ECE = \frac{1}{|n|} \sum^n_{i=1} | C(B_i) - Acc(B_i)| \  |B_i|
\end{equation}

In \cite{nixon2019measuring} they evaluate this metric and highlight potential problems when using it.
In particular it is shown that many somewhat arbitrary choices in the metric such as the choice of norm, binning strategy, and class weighting can produce metrics that yield slightly different measures of calibration error. 
The authors produce a total of 32 possible variations of this calibration error and when comparing calibration methods the ranking is generally inconsistent across these 32 metrics. 

It is likely that the optimal calibration metric will depend upon the use case in mind, therefore general comparisons between calibration methods are likely to be a difficult problem. 

In this work we evaluate calibration methods using the probability that the classifier is correct together with adaptive binning schemes and weighted by the total number of points in each bin rather than also conditioning on the class - this corresponds to the general calibration metric 20 defined in the appendix of \cite{nixon2019measuring}. 
We chose this metric for ease of comparison with other studies (such as \cite{GuoCalibrationModernNeural2017, naeini2015obtaining}) and because the metric is sufficient to highlight the utility of MACEst for confidence estimation. 

As noted previously, in particular in \cite{nixon2019measuring}, it has been shown that calibration errors can sometimes be misleading. 
It is therefore important to have additional criteria against which to measure estimators. 
The quality of a set of probabilistic predictions has been a long-standing issue in many fields such as Meteorology where, as well as calibration metrics, the notion of \emph{proper} scoring rules have developed. 

Proper scoring rules are metrics which, when minimised, correspond to approximating the ground truth probability distribution \citep{gneiting2007strictly}. 
This means that a \emph{biased} forecaster will perform worse in these metrics than an honest forecaster - evaluation by proper scoring rules was advocated by \cite{lakshminarayanan2017simple}. 
The two most popular scoring rules are the Brier Score (\cite{murphy1977reliability})  and the Negative log loss (NLL) (\cite{friedman2001elements}). 

As we are looking to draw parallels between trust scores and calibration methods we are trying to estimate the probability that our point prediction is correct, therefore these scoring rules are also defined with respect to a prediction being correct, i.e. $O_i =  \large{\delta_{\hat{y_i}, y_i}} $. 
This is equivalent to giving class probabilities in the binary case but equates to a one vs. rest strategy in multi-class problems. 
We therefore use the Brier Loss as follows:
\begin{equation}
    BL = \sum^n_{i=1} (p_i - O_i)^2
\end{equation}

And the Negative Log Likelihood (NLL) as:
\begin{equation}
    NLL = - \sum^n_{i=1} {(O_i\log(p_i) + (1 - O_i)\log(1 - p_i))}
\end{equation}

Intuitively both of these metrics penalise (via squared loss for Brier and logarithmic loss for NLL) incorrect predictions with high confidence but also correct predictions with low confidence. 
Therefore performing well in these metrics generally corresponds to predicting correctly with high confidence and predicting incorrectly with low confidence. 
This differs slightly with respect to calibration metrics where the only measured quantity is how well the predicted confidence scores approximate the empirical accuracy of the predictions.

All evaluation metrics are evaluated, as will be shown below, using K-fold cross-validation by taking the error intervals on these estimates to be twice the standard deviation of the K folds - thereby approximating roughly the 95\% error interval.
This provides a useful way to evaluate if there are any \emph{significant} differences between calibration methods. 

\section{Related Work}

\subsection{Confidence Calibration} \label{sec:Calibration}
Confidence calibration was first studied in \cite{platt1999probabilistic}, since then there have been many more calibration techniques developed \citep{ZadroznyTransformingClassifierScores2002, Niculescu-MizilPredictinggoodprobabilities2005, naeini2015obtaining, GuoCalibrationModernNeural2017, kull_beyond_2019}. 
Each of these techniques aim to transform the raw scores from the point prediction algorithms into a confidence score that approximates the empirically derived accuracy. 
This is generally by some distributional transformation or scaling, where the parameters required to do the transformation are learnt using a single hold out set of data. 

For example if the point prediction model predicts a confidence score of 80\% one of the techniques above may learn a transformation that down-scales that score to a 70\%. 
It should be noted here that these transformations are dependent upon the point prediction algorithm score only: they are not explicitly dependent upon a given data point for which they are asked to give a confidence estimate.

These methods have been shown to work very well across a range of machine learning problems. 
However, as we will show in section \ref{sec: not trustworthy}, they learn a \emph{global representation} of uncertainty and are therefore often only actually learning to model aleatoric uncertainty. 
This means that they are vulnerable to giving overly confident predictions due to ignoring epistemic uncertainty. 

\subsection{Trust Scores}
In (\cite{JiangTrustNotTrust2018}) the authors introduce a method that produces a \emph{trust score} for model predictions, i.e. a high trust score means that a model is likely to be correct and vice versa. 
This notion is very closely related to the confidence one has that a prediction is correct. 
The problem of trust is approached by looking at the similarity between the point you want to predict and the training data.
The relative distances between a set of $k$ nearest neighbours from the predicted class and from all other classes are then compared. 
For example a trust score of $1$ corresponds to the prediction being equally as close to the $k$ nearest neighbours from the predicted class to those of all other classes; a trust score of $2$ means it is twice as close to the predicted class, etc. 
The intuition here is that if a data point is similar to previously seen points of the same class then the prediction is more likely to be correct and therefore more \emph{trustworthy}.  
This can be interpreted as estimating the epistemic uncertainty: when the trust score is large, the distance to similar points is small and therefore the epistemic uncertainty is small. 
One would then expect that there is a greater chance of the prediction being correct, i.e. more likely to be trustworthy. 

\subsection{Other related works}

Quantifying the uncertainty in a prediction is most naturally considered within a Bayesian framework, parametric methods such as Bayesian parametric modelling (\cite{tipping2003bayesian} \cite{congdon2014applied}) and non-parametric methods such as Gaussian Process (GP) Regression (\cite{rasmussen2003gaussian, williams2006gaussian}) and Bayesian Neural Networks (\cite{neal2012bayesian}) have proven to be effective for providing prediction intervals. 
These methods however suffer from some drawbacks that MACEst attempts to address, generally they are considerably more computationally expensive at both training and inference time than algorithms that do not include confidence estimates: this is potentially a major barrier for many applications. 

There is in general no method that the authors are aware of to explicitly combine Bayesian uncertainty estimates with an arbitrary point prediction algorithm. 
MACEst is not a Bayesian algorithm however, and, as described in the introduction, it is motivated by some of the underlying Bayesian principles. 
It seeks to bridge the gap by combining some of the benefits of Bayesian modelling with less computational costs and, by being compatible with any point prediction algorithm, considerably more flexibility.

Outside of the explicitly Bayesian framework there are several methods that have been utilised for the problem of confidence and uncertainty estimation. 
The Dropout method introduced in \cite{gal_dropout_2015} could potentially be applied to other methods by perturbing or dropping model parameters and generating a range of predictions, however such Monte Carlo simulation methods used to get a good coverage of the parameter space where the number of model parameters is large are computationally expensive.
Ensemble methods similar to those introduced for neural networks in \cite{lakshminarayanan2017simple} could also be applied for any model, however the cost of training a large enough ensemble to is often prohibitive for many applications.

\section{Calibrated Models Are Not Necessarily Trustworthy } \label{sec: not trustworthy}
Above we have seen that there are two different approaches to answering very similar questions, i.e. how confident are we in a prediction (calibration methods)? 
And how much can we trust a prediction (Trust scores)? 
Intuitively these seem to be very related, yet the two approaches estimate their score very differently. 
This section asks the question: can we trust well calibrated estimators? 

To answer this question we perform classification on the classic MNIST data set (\cite{lecun2010mnist}), which consists of $60,000$ examples of 28x28 pixel images of handwritten digits with the task of classifying those digits. 
As trust score uses a distance metric the raw pixels are not suitable, so we took $50$ principal components and used those as our features. 
We then trained a simple random forest model as the point prediction model and used a selection of the cited calibration methods described above to compute confidence estimates. 
We now look at the correlation between trust scores and these confidence estimates. 

\begin{figure}[t]
    \centering 
    \includegraphics[width= 0.9 \textwidth]{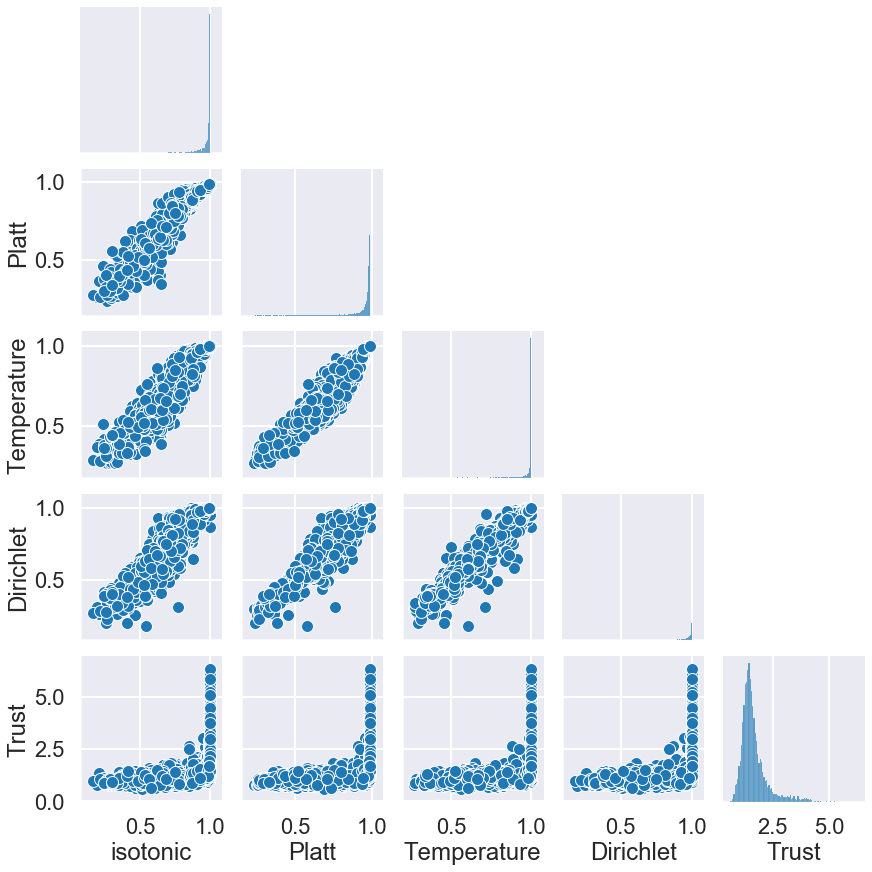}
    \caption{Pair plots comparing trust scores and confidence predictions when predicting on an unseen test set for the mnist dataset. Trust scores are correlated strongly with confidence predictions here
    }
    \label{fig:conf_vs_trust_test}
\end{figure}

Figure \ref{fig:conf_vs_trust_test} shows the joint distributions for the confidence estimates and trust scores evaluated on an unseen test data set. 
The point prediction model is generally very accurate ($\approx 95 \% $) and thus, as expected, for each confidence estimator we have a confidence distribution which is very high, thereby corresponding to high ($ \approx 2$) trust scores. 
We calculated the Spearman rank correlation coefficient between the confidence predictions and trust scores, and found that for each calibration model the correlation between them is very strong (Isotonic = $0.797$, Platt = $0.803$, Temp = $0.814$, Dirichlet= $0.773$). 
This is generally what we should expect: the models are reliable and predicting well and so both the confidence and trust scores are high. 
In this situation we see that the standard calibration methods are working well, and the confidence estimates are reliable.

\begin{figure*}[t] 
        \includegraphics[width= 0.3\textwidth]{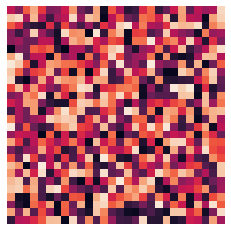}
        \centering
        \caption{An example of the uniform random noise which we asked our model (which was trained on MNIST) to classify. 
        Confidence estimates calibrated using Dirichlet, Temperature, Isotonic and Platt all reported $>85\%$ confidence.}
        \label{fig:mnist_noise}
        \includegraphics[width= 0.8\textwidth]{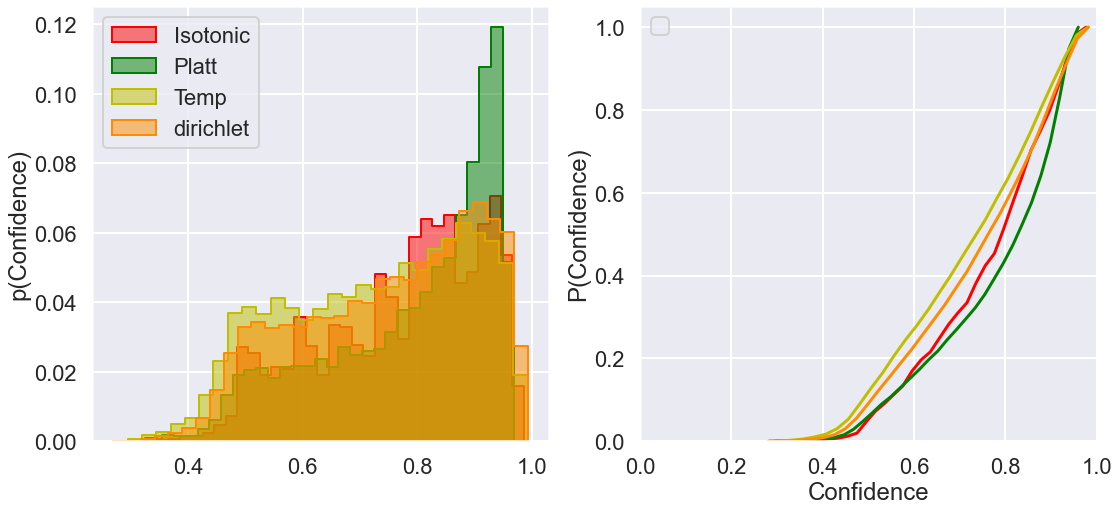}
        \centering
        \caption{Example of $10^4$ confidence estimates on random noise similar to Figure \ref{fig:mnist_noise}: we show the probability distribution function (left) and the smoothed cumulative density function (right). 
        These distributions \emph{should} be shifted towards $0.1$ thereby indicating the model's lack of confidence given the presence of pure noise: instead, we see for all calibration methods that there are a large number of high confidence predictions despite the input.}
        \label{fig:mnist_noise_dist}
\end{figure*}

Next, we simulated 28x28 pixel images using $10,000$ different uniform noise realisations and asked our model to make predictions on this data (See figure \ref{fig:mnist_noise} for an example of a noise realisation) after having trained the models on the standard MNIST dataset.
As this does not look like any digit, we would expect the confidence estimates also be very low for any prediction and should, ideally, return something like a uniform distribution across all classes that would then reflect minimal confidence.  
We can see from Figure \ref{fig:mnist_noise_dist} that although the confidence is considerably lower than most of the in-sample confidence estimates, it is still often high. 
Most predictions are greater than $0.5$ and there are a considerable number ($\sim$ 30\%) which are greater than $0.8$. 

\begin{figure}[t]
    \centering 
    \includegraphics[width= 0.9 \textwidth]{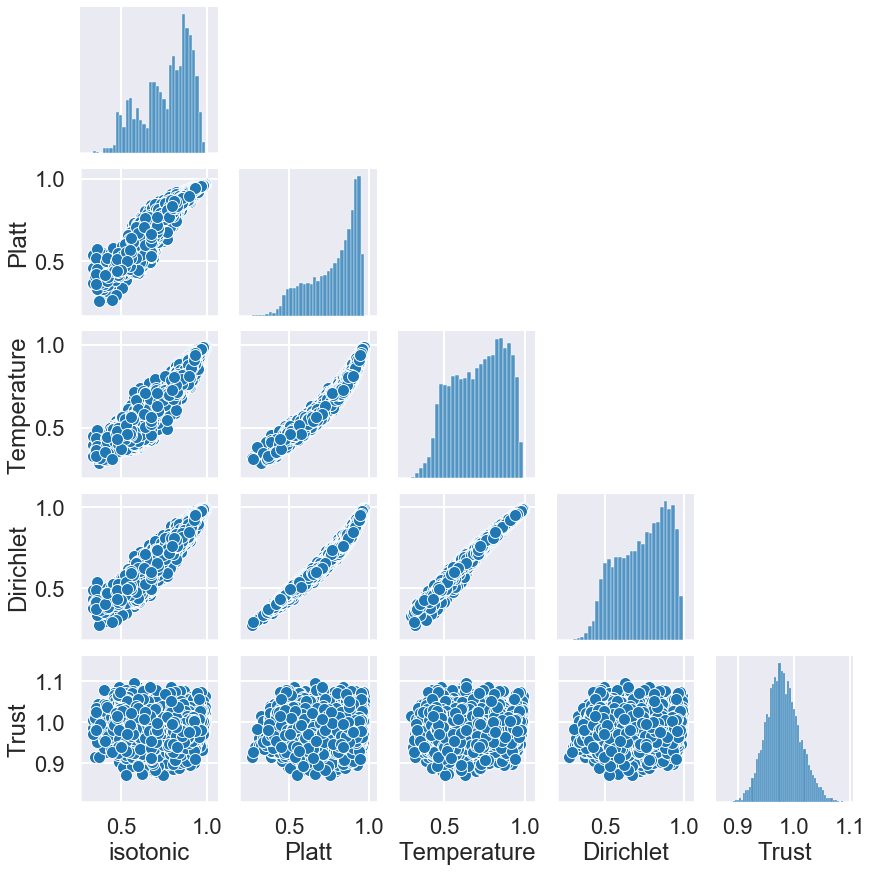}
    \caption{Pair plots comparing trust scores and confidence predictions when predicting uniform noise: when epistemic uncertainty is large then trust scores and confidence estimates become uncorrelated. 
    Trust scores are generally low $\sim 1$ but confidence predictions can be high. 
    Note the correlation between the predictions of the models, this is because they all perform a slightly different transformation on the same set of original predictions: therefore the ranking of points will not change.}
    \label{fig:conf_vs_trust_noise}
\end{figure}

What about the trust scores? 
We see in  Figure \ref{fig:conf_vs_trust_noise} the trust scores are now much lower. 
95\% of trust scores are now $\leq 1.05$, this reflects our intuition that noise should be roughly as close to the predicted class as any other. 
When comparing to the confidence scores however we see that the correlation between trust scores and confidence estimates has effectively disappeared  (Isotonic = $0.056$ , Platt = $0.085$, Temp = $0.108$, Dirichlet= $0.089$).  
This example highlights that despite confidence estimators being calibrated to the data they are not necessarily trustworthy, as posited by \cite{JiangTrustNotTrust2018}.

\begin{figure}[t] 
    \includegraphics[width= 1.0 \textwidth]{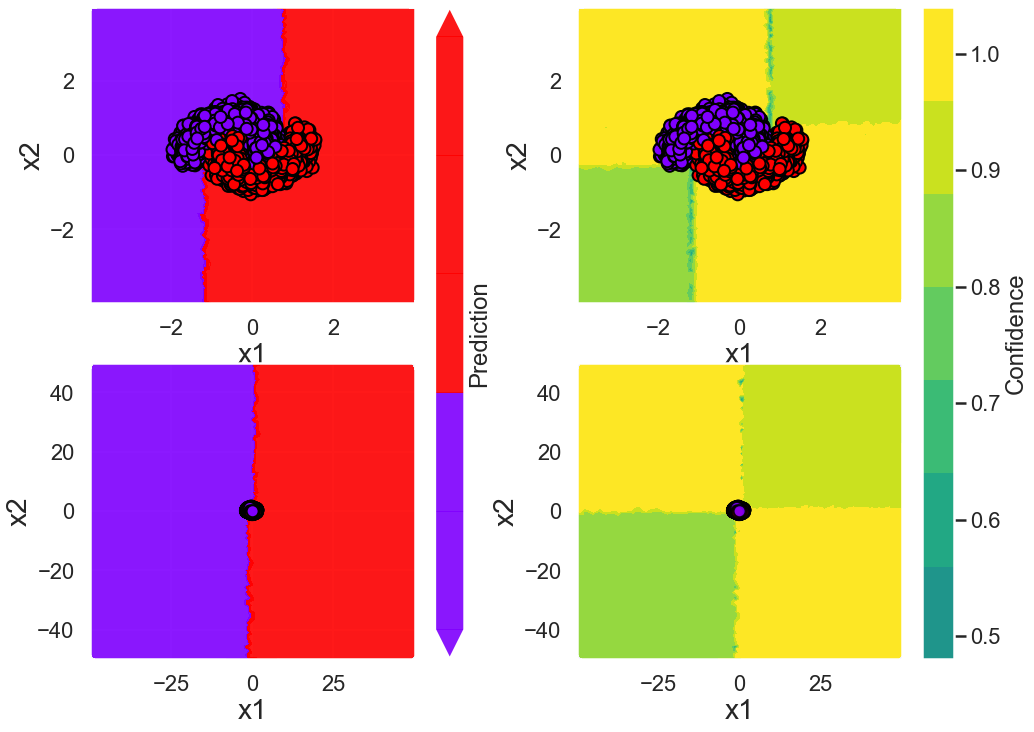}
    \centering
    \caption{Example illustrating the problem with extrapolating confidence: because the confidence estimate does not account for the \emph{similarity} to the training data, the model confidence will not decrease however far we extrapolate from the data from which the model learnt.}
    \label{fig:extrapolating_conf}
\end{figure}

We argue that this effect is fundamental to the way classification models work in practice.
The classification paradigm is effectively to learn a global decision boundary and split the features into distinct regions of classes. 
This is a natural and incredibly effective way to perform classification, however we argue it is not a good way to estimate confidence. 
This is because it does not take into account the \emph{similarity} of a given point to the data that the model used to train, resulting in epistemic uncertainty never being accounted for.
If a point falls far away from the decision boundary it is confidently predicting a label even if it is completely different to any of the examples that trained a model (\cite{gal_dropout_2015}). 
This can be seen clearly in the simple example shown in Fig. \ref{fig:extrapolating_conf}: here we have an example with a set of simulated points in two classes that follow a spiral.
There is a good distinction between the classes and therefore the model is able to clearly split the feature space into regions where it will predict either red and blue. 
In the regions close to the data the confidence is very high because the model is generally predicting very accurately. 
As we move away from the training data, we see that the predictions are extrapolated according to the global boundary which splits the feature space. 
We also see however that the confidence estimates are extrapolated, this means that in regions of the feature space very far away from the training data the model is still returning very high confidence results. 
We argue that the confidence estimate should \emph{decrease} as we move away from the training data because the epistemic uncertainty is considerably higher in these regions.

Both of these examples highlight a fundamental problem with using the model estimates as a starting point for a confidence estimate. 
Confidence estimates derived from model estimates can generally be calibrated to estimate the aleatoric uncertainty well, however confidence estimates are not reliable when being used for data that is \emph{significantly} different from the training set. 
This can often result in over-confident predictions, despite a high degree of epistemic uncertainty being present. 

Trust scores, as an estimate of the distance to similar data from the training set, are an effective way of highlighting cases where the epistemic uncertainty is high, however they do not have any mechanism to account for aleatoric uncertainty and are also not easy to interpret as a likelihood or probability of being correct.
The algorithm we now present seeks to bridge the gap between these two approaches by producing well calibrated confidence estimates which account for both epistemic and aleatoric uncertainty.

\section{MACEst}
We start with the heuristic that confidence is a local quantity, e.g. if the model is terrible globally then there are still some predictions for which the model can be very confident. 
Similarly if the model is very accurate (even perfect) globally on a given test set then for any given new prediction there exists the possibility that the model is not capable of providing a good answer and should therefore have very low confidence for this point. 
We also argue above that the point prediction model may fail to produce good confidence estimates and therefore the confidence for a prediction should be estimated independently of the model that produced the classification.

The following assumptions outline our approach to model the local confidence estimate:
\begin{enumerate}
    \itemsep-0.55em 
    \item A point that not similar to any training data is an unusual point and predictions on that point should not have a high confidence value associated with them.
    \item The distance to a set of nearest neighbours from the training data is a good proxy for how similar any point is to the training data, and as a consequence this can then be used to estimate the epistemic uncertainty.
    \item The accuracy of a set of predictions for a set of $k$ nearest neighbours can be used as a proxy for the local prediction variance, i.e. the aleatoric uncertainty for the prediction can be estimated by the accuracy in these $k$ nearest neighbours.
\end{enumerate}

 To implement this for a given prediction on a point $\mathbf{x_*}$, we define the local neighbourhood by finding the $k$ nearest neighbours to $\mathbf{x_*}$. 
 Using these nearest neighbours we then attempt to directly estimate proxies for the epistemic and aleatoric uncertainty for predictions in that neighbourhood. 
 Once we have estimated these quantities we define a simple parametric function of them and calibrate the function so that our confidence estimates approximate the empirical accuracy. 
 By modelling these two effects directly, confidence estimates produced by MACEst are able to encapsulate the local variance accurately whilst also being aware of when the model is being asked to predict a point that is very different to those it has been trained on. 
 This makes MACEst robust to problems such as overconfident extrapolations and bad out of sample predictions.
 We now describe the algorithm in more detail.

\subsection{Algorithm}
MACEst requires that a given dataset is split into four distinct sets: the standard labelled training set for the point prediction model; a set of labelled data from which MACEst finds the nearest neighbours for any given point;\footnote{One could use the point prediction training data for this however this may mean that there is bias when estimating the local prediction accuracy due to information leakage when training the point prediction model. We therefore use an extra split in our dataset to avoid this.} a set of labelled data to optimise the MACEst parameters (see equation \ref{eq:mace} below) by minimising the expected calibration error; and finally the unseen hold-out data used to test the model, one can use the same test data for both MACEst and the point prediction model.

Next, one must define a notion of similarity, this is equivalent to defining a sensible distance measure between points. 
Often choices such as a Euclidean distance will be sufficient if the co-ordinates are sensible, i.e. not highly correlated, however any measure of similarity may be suitable depending upon the data one is trying to model. 
For difficult problems one can generally improve the performance of MACEst by exploiting techniques such as PCA, Word2Vec, transformers, ect (\cite{jolliffe2003principal, hinton2006reducing, mikolov2013distributed, devlin2018bert}) to embed the data into a space where the metric better reflects the similarity between points.
In a standard pipeline this would be a modular step, i.e. the downstream algorithm works with any definition of similarity so one can swap in any embedding technique depending upon the problem. 
The calibration of MACEst may then be better or worse depending upon the choice of metric.

Once we have defined a notion of similarity, for any given point $\mathbf{x_*}$, we find the $k$ \emph{most similar} points from the known MACEst training data. 
Finding the $k$ nearest neighbours exactly has $O(n^2)$ complexity which is prohibitively expensive for many situations. 
Finding Approximate Nearest Neighbours (ANN) can however be considerably less expensive and is an active area of research.
In this work we used the Hierarchical Navigable Small World algorithm \citep{malkov2018efficient} for ANN search, which leads to a complexity of $O(n\log n)$ during training in order to build the graph and then $O(\log n)$ complexity at inference once the graph has been built. 
Due to the simplicity of MACEst these are usually the dominant costs for both training and inference.

We note here that the nearest neighbour search is also a modular step. 
MACEst, in principle, does not rely on any particular nearest neighbour algorithm so other methods for example \cite{JDH17, avq_2020} may be more appropriate depending on the application. 
See \cite{aumuller2017ann} for a comparison of approximate nearest neighbour methods.

Once we have found the $k$ nearest neighbours we use these $k$ points to estimate proxies for both epistemic and aleatoric uncertainty. 
The local aleatoric uncertainty will be the (distance weighted) error, i.e. the number of incorrect predictions on these $k$ points each weighted by the distance from the point where we would like to predict.
The epistemic uncertainty will be approximated by the average (weighted by the rank order of closeness) distance to these $k$ neighbours. 
We define the error as $\epsilon(\mathbf{x_k})$ and the average distance to the set of neighbours as $\mathcal{K}(\mathbf{x_*}, \mathbf{x_k}) $. 
The un-normalised confidence score is then approximated to be a simple function of these terms as follows:
\begin{equation}\label{eq:mace}
    \sigma_* = \alpha\epsilon(\mathbf{x_*}, \mathbf{x_k}) + \beta \mathcal{K}(\mathbf{x_*},\mathbf{x_k})
\end{equation}

Where we define $\epsilon(\mathbf{x_k})$  and $\mathcal{K}(\mathbf{x_*}, \mathbf{x_k}) $\footnote{In the average distance definition below, we weight the distance by the rank order in terms of closeness, we found that this heuristic gave a more robust average in cases where the $k$ neighbours were split between several disjoint clusters and therefore the average distance is not well-defined.} as:
\begin{equation}
   \mathcal{K}(\mathbf{x_*},\mathbf{x_k}) = \sum_{i=i}^k \frac{d(\mathbf{x_*},\mathbf{x_i})}{i}
\end{equation}

And,
\begin{equation}
    \epsilon(\mathbf{x_*}, \mathbf{x_k}) = \sum_{i=i}^k \frac{\Large{\delta_{\hat{y_i}, y_i}}}{d(\mathbf{x_*},\mathbf{x_i})}
\end{equation}

In both definitions, we assume that the $k$ points are ordered by their relative distances to $x_*$, and we used the Euclidean distance\footnote{The Euclidean distance may be calculated in an embedded space rather than the Euclidean distance between the data points themselves.} as our distance function:
\begin{equation}
   d(\mathbf{p},\mathbf{q}) = \sqrt{\sum_{j=1}^D (p_j - q_j)^2}
\end{equation} 

We can then see that if points are very similar to the training set then $ \beta \mathcal{K}(\mathbf{x_*},\mathbf{x_k}) \ll \alpha\epsilon(\mathbf{x_k}) $ and the uncertainty will be dominated by the local aleatoric uncertainty but if points are very distant then $ \beta \mathcal{K}(\mathbf{x_*},\mathbf{x_k}) \gg \alpha\epsilon(\mathbf{x_k}) $ and the epistemic uncertainty will dominate. 
The coefficients $\alpha $ and $ \beta $ can then be interpreted by the relative importance of both factors and can be learnt during the MACEst training phase. 

For the classification problem we repeat these steps for each class and then have a $\sigma_*$ for each class.
These scores are then normalised first by dividing by the average score across all classes for a given point: this can be interpreted as returning a score for each class which is relative to the average uncertainty across all classes. 
We then apply a negative softmax normalisation to these relative scores in order to return probabilities within the interval $[0,1]$.

Note that for the purposes of this work we are only concerned with estimating the probability that the point prediction produced by the original model is correct, we therefore minimise the expected calibration error for the predicted class only when training. 
This need not necessarily be the largest confidence according to MACEst because the point prediction model and MACEst are calculating the relative likelihood of each class independently. 
We found empirically that this rarely happens but in cases where they clearly disagree this suggests that either a point is particularly unusual, and therefore the point prediction algorithm is likely to perform badly on them, or the distance metric used by MACEst may not be optimal. 

In principle this algorithm could of course be extended to be made more flexible by adding non-linearities or including higher order effects for each factor however, in principle, we found that a simple linear model was generally sufficient for the experiments shown below. 

\section{Experiments}
We evaluated MACEst with three different experiments, in each case comparing MACEst to other standard calibration methods. 
The first experiment looked at the performance of MACEst compared to other methods on several standard UCI datasets. 
The models were compared on a held-out test dataset where this test set was unseen but drawn from the same dataset which the models were trained on: this experiment therefore focus on how the models deal with aleatoric uncertainty. 
The second experiment simulated a situation where data changes after a point prediction model has been trained: here, standard confidence estimators may fail thereby highlighting a key difference between the way MACEst and other methods deal with epistemic uncertainty. 
Finally we repeated the experiment shown in section \ref{sec: not trustworthy} with MACEst to show that MACEst is generally robust to problems caused by predicting on out-of-sample data, we also show that the relationship between Trust scores and MACEst remains highly correlated with both in-sample and out-of-sample data.

\subsection{Aleatoric Experiments}
For this set of experiments we used seven datasets from the UCI collection (\cite{Dua:2019}). 
As discussed above, the choice of similarity metric will influence the results so for all datasets we first performed a simple embedding strategy: Principal Component Analysis for non-text datasets and TF-IDf followed by Neighbourhood Component analysis \citep{goldberger2004neighbourhood} for text datasets. 
Once we had embedded the data we measured similarity using the Euclidean distance in this space. 
We then trained a Random Forest classifier \citep{breiman2001random} and calibrated the Random Forest uncertainty using each of the MACEst, Platt, Isotonic, Temperature scaling and Dirichlet Calibration methods.
Finally we computed the ECE, NLL, and Brier score using K-fold cross validation with $10$ folds. 
We then reported the 95\% error intervals as twice the standard deviation over these $10$ folds.

Looking first at Tables \ref{tab:nll} and \ref{tab:brier}, where we evaluated the models using the proper scoring rules Negative Log Loss and Brier loss, we see that MACEst records the lowest mean Negative Log Likelihood (NLL) for two datasets MNIST and EEG. 
For these two datasets we see a significant difference between MACEst and the other models. 
The large difference in negative log likelihood is likely because MACEst is able to reduce the confidence of \emph{bad} predictions and therefore will have less high confidence predictions that are incorrect. 
This increases the \emph{resolution} of the model as it is able to distinguish between very high and very low confidence predictions to a greater degree.
We see that rankings are broadly consistent across the different metrics however there are some differences.
For example on the Letters dataset when looking at the mean score MACEst is ranked fourth on NLL but second on Brier loss.

Looking now at table \ref{tab:ece} we see similar results, this is expected as calibration is one part of the proper scoring rules. 
Again we see some differences between the rankings of various models on each of the datasets. 
MACEst is again clearly the lowest on MNIST and EEG with the rankings varying across other datasets. 
In general across each of the metrics MACEst seems to perform less well on the NLP tasks, this is not surprising as the induced similarity metric used in this work is very simple. 
It is likely that by using more sophisticated text embedding methods we would be able to perform better on these datasets.

To summarise the results, we find that, as has been seen previously, the differences between various methods are very problem dependent, and that the ordering of methods changes across the experiments as well as across different metrics. 
Looking at the error intervals across the three metrics we see that often the distributions are overlapping meaning that it is difficult to draw statistically significant conclusions regarding a \emph{best} model on any of the datasets. 
We do however see that MACEst is generally competitive with other methods commonly used to tackle the problem of confidence estimation. 

It is likely that for many problems each of the calibration methods generally performs well and the differences between them are small. 
We therefore suggest that for problems where the data is unlikely to differ considerably to the training data then each of the calibration methods will likely be \emph{good enough} and that ultimately the choice of method should be based upon other problem-dependent factors such as computational costs, the specific objective function, and whether epistemic uncertainty is likely to be a problem. 

\begin{table}[]
\centering
\resizebox{\textwidth}{!}{%
\begin{tabular}{l|lllll}
\cline{2-6}
 & \multicolumn{1}{l|}{\textbf{MACEst}} & \multicolumn{1}{l|}{\textbf{Platt}} & \multicolumn{1}{l|}{\textbf{Isotonic}} & \multicolumn{1}{l|}{\textbf{Temperature}} & \multicolumn{1}{l|}{\textbf{Dirichlet}} \\ \hline
\multicolumn{1}{|l|}{\textbf{EEG}} & $ 0.147 \pm 0.06 $ & $0.217 \pm 0.05$ & $0.228 \pm 0.06 $ & $0.225 \pm 0.06 $ & $0.221 \pm 0.06 $ \\ \cline{1-1}
\multicolumn{1}{|l|}{\textbf{Text Sentiment}} & $ 0.525 \pm 0.26 $ & $ 0.49 \pm 0.17$ & $ 0.504 \pm 0.26 $ & $ 0.520 \pm 0.26 $ & $ 0.607 \pm 0.61$ \\ \cline{1-1}
\multicolumn{1}{|l|}{\textbf{Letters}} & $ 0.199 \pm 0.03 $ & $0.20 \pm 0.02$ & $ 0.187 \pm 0.17 $ & $ 0.168 \pm 0.02 $ & $ 0.170 \pm 0.04$ \\ \cline{1-1}
\multicolumn{1}{|l|}{\textbf{Mnist}} & $ 0.07 \pm 0.01  $ & $0.13 \pm 0.01 $ &  $ 0.125 \pm 0.01  $ &  $ 0.118 \pm 0.01  $ &  $ 0.115 \pm 0.01  $ \\ \cline{1-1}
\multicolumn{1}{|l|}{\textbf{Fashion}} & $ 0.30 \pm 0.01  $ & $0.30 \pm 0.01$ & $ 0.28 \pm 0.01  $ & $ 0.28 \pm 0.01  $ & $ 0.28 \pm 0.01  $ \\ \cline{1-1}
\multicolumn{1}{|l|}{\textbf{20 news}} & $ 0.38 \pm 0.08  $ & $0.29 \pm 0.08$ &$ 0.38 \pm 0.26  $  & $ 0.10 \pm 0.27  $ &$ 0.28 \pm 0.14  $  \\ \cline{1-1}
\multicolumn{1}{|l|}{\textbf{Adult}} & $ 0.43 \pm 0.02  $ & $0.42 \pm 0.02$ & $ 0.40 \pm 0.02  $ & $ 0.49 \pm 0.07  $ & $ 0.49 \pm 0.07  $ \\ \cline{1-1}
\end{tabular}%
}
\caption{Negative log likelihoods calculated for each of the benchmark UCI datasets, error bars are estimated using K-fold validation with 10 folds.
}
\label{tab:nll}
\end{table}

\begin{table}[]
\centering
\resizebox{\textwidth}{!}{%
\begin{tabular}{l|lllll}
\cline{2-6}
 & \multicolumn{1}{l|}{\textbf{MACEst}} & \multicolumn{1}{l|}{\textbf{Platt}} & \multicolumn{1}{l|}{\textbf{Isotonic}} & \multicolumn{1}{l|}{\textbf{Temperature}} & \multicolumn{1}{l|}{\textbf{Dirichlet}} \\ \hline
\multicolumn{1}{|l|}{\textbf{EEG}} & $ 0.042 \pm 0.02 $ &  $0.065 \pm 0.02 $ & $0.068 \pm 0.02 $ & $0.068 \pm 0.02 $  & $ 0.067 \pm 0.02 $ \\ \cline{1-1}
\multicolumn{1}{|l|}{\textbf{Text Sentiment}} & $ 0.162 \pm 0.07 $ & $0.160 \pm 0.07 $ & $ 0.156 \pm 0.07 $ & $ 0.159 \pm 0.08 $ & $ 0.167 \pm 0.06$ \\ \cline{1-1}
\multicolumn{1}{|l|}{\textbf{Letters}} & $ 0.054 \pm 0.01 $ & $0.055 \pm 0.007$ & $ 0.057 \pm 0.01 $ & $ 0.052 \pm 0.01 $ & $ 0.055 \pm 0.01$ \\ \cline{1-1}
\multicolumn{1}{|l|}{\textbf{Mnist}} &  $ 0.02 \pm 0.003  $ & $0.037 \pm 0.003$ &  $ 0.037 \pm 0.002  $ &  $ 0.035 \pm 0.003  $ &  $ 0.035 \pm 0.003  $ \\ \cline{1-1}
\multicolumn{1}{|l|}{\textbf{Fashion}} & $ 0.095 \pm 0.005  $ & $0.09 \pm 0.005 $ & $ 0.089 \pm 0.004  $ &$ 0.0089 \pm 0.03  $  & $ 0.089 \pm 0.03  $ \\ \cline{1-1}
\multicolumn{1}{|l|}{\textbf{20 news}} & $ 0.109 \pm 0.02  $ & $0.083 \pm 0.03$ & $ 0.079 \pm 0.03  $ & $ 0.079 \pm 0.03  $ & $ 0.082 \pm 0.03  $ \\ \cline{1-1}
\multicolumn{1}{|l|}{\textbf{Adult}} & $ 0.13 \pm 0.007  $ & $ 0.13 \pm 0.006$ & $ 0.13 \pm 0.006  $ & $ 0.13 \pm 0.005  $ & $ 0.13 \pm 0.005  $ \\ \cline{1-1}
\end{tabular}%
}
\caption{Brier score calculated for each of the benchmark UCI datasets, error bars are estimated using K-fold validation with 10 folds}
\label{tab:brier}
\end{table}

\begin{table}[]
\centering
\resizebox{\textwidth}{!}{%
\begin{tabular}{l|lllll}
\cline{2-6}
 & \multicolumn{1}{l|}{\textbf{MACEst}} & \multicolumn{1}{l|}{\textbf{Platt}} & \multicolumn{1}{l|}{\textbf{Isotonic}} & \multicolumn{1}{l|}{\textbf{Temperature}} & \multicolumn{1}{l|}{\textbf{Dirichlet}} \\ \hline
\multicolumn{1}{|l|}{\textbf{EEG}} & $ 0.017 \pm 0.05 $ & $0.022 \pm 0.006$ & $ 0.0222 \pm 0.011 $ & $0.023 \pm 0.010 $ & $0.24 \pm 0.008 $ \\ \cline{1-1}
\multicolumn{1}{|l|}{\textbf{Text Sentiment}} & $ 0.137 \pm 0.05 $ & $0.123 \pm 0.04 $ & $ 0.104 \pm 0.05 $ & $ 0.125 \pm 0.05 $ & $ 0.21 \pm 0.03$ \\ \cline{1-1}
\multicolumn{1}{|l|}{\textbf{Letters}} & $ 0.03 \pm 0.01 $ & $0.05 \pm 0.01 $ & $ 0.04 \pm 0.01 $ & $ 0.02 \pm 0.01 $ & $ 0.02 \pm 0.01$ \\ \cline{1-1}
\multicolumn{1}{|l|}{\textbf{Mnist}} &  $ 0.005 \pm 0.002  $ & $0.020 \pm 0.003$ &  $ 0.016 \pm 0.005  $ &  $ 0.007 \pm 0.002  $ &  $ 0.007 \pm 0.03  $ \\ \cline{1-1}
\multicolumn{1}{|l|}{\textbf{Fashion}} & $ 0.019 \pm 0.08  $ & $0.40 \pm 0.005$ & $ 0.018 \pm 0.003  $ & $ 0.013 \pm 0.004  $ & $ 0.011 \pm 0.003  $ \\ \cline{1-1}
\multicolumn{1}{|l|}{\textbf{20 news}} & $ 0.051 \pm 0.02  $ & $0.068 \pm 0.022$ & $ 0.029 \pm 0.01  $ & $ 0.032 \pm 0.01  $ & $ 0.035 \pm 0.02  $ \\ \cline{1-1}
\multicolumn{1}{|l|}{\textbf{Adult}} & $ 0.045 \pm 0.01  $ & $0.058 \pm 0.006 $ &$ 0.022 \pm 0.005  $  &$ 0.024 \pm 0.06  $  & $ 0.026 \pm 0.08  $ \\ \cline{1-1}
\end{tabular}%
}
\caption{ECE calculated for each of the benchmark UCI datasets, error bars are estimated using K-fold validation with 10 folds}

\label{tab:ece}
\end{table}

\subsection{Epistemic Experiment} \label{sec: epistemic}
Motivated by use cases where models are trained at a fixed moment in time (e.g. each morning, start of each week) and then used for some period of time after, we split the data into training and testing: a model was then trained on the former while the latter set had noise artificially added to its features, thereby simulating heteroskedastic noise, i.e. replicating the situation where data often changes throughout the life cycle of a model. 
In these cases it is therefore important that a model is able to understand that the data is changing and adjust the confidence in predictions accordingly.
To add simulate noise, we consecutively added Gaussian noise to the features in the test data by increasing the standard deviation of the noise at each iteration. 
We then recorded the average point prediction accuracy and the mean confidence for each calibration model.

\begin{figure*} [t]
        \centering
        \includegraphics[width= 0.75\textwidth]{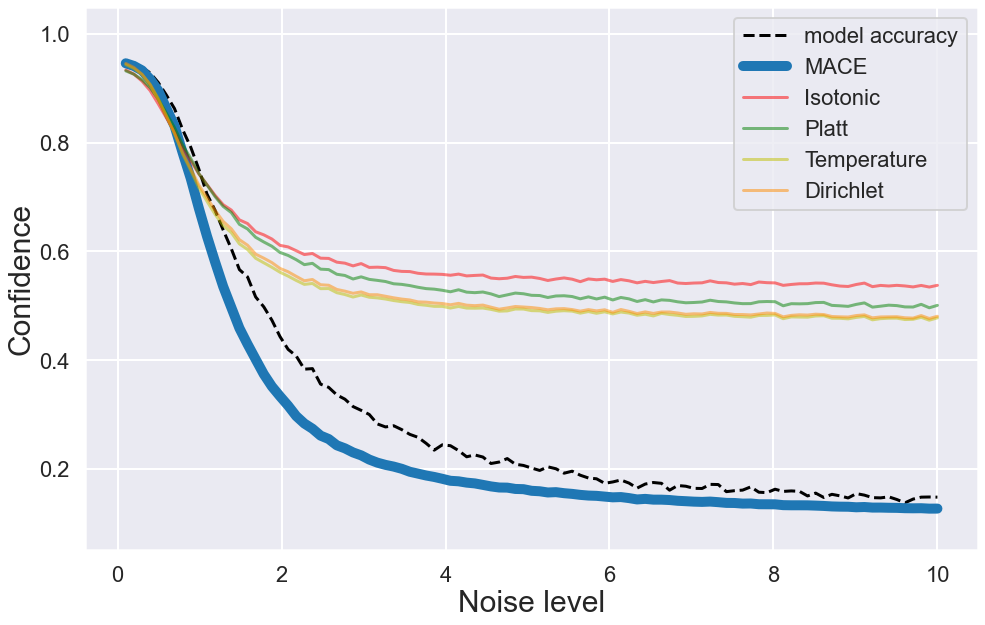}
        \caption{Here we add zero mean Gaussian noise with a standard deviation defined by the noise level.
        This simulates data drift by iteratively making the test set more different to the training data. 
        We show the mean of the test set confidence distribution for each model and see that because MACEst explicitly calculates the epistemic uncertainty it is able to track the degradation in model performance considerably better than the other calibration methods.}
        \label{fig:epi_plot}
\end{figure*}

The results in Figure \ref{fig:epi_plot} show the average accuracy of the model and the average confidence estimate of each confidence estimator. 
We see that initially all models respond to the noise in a very similar way, i.e. dropping the mean confidence to follow the trend in mean prediction accuracy. 
As the amount of noise increases the model accuracy continues to decrease, until eventually it is scoring around 10\% which corresponds to a random choice. 
When we compare the accuracy to the confidence estimates we see that each of the models, apart from MACEst, decreases its confidence scores but are bounded at around 50\% despite the continuous decrease in model performance. 
This is likely because the other calibration models, that have learnt a fixed transformation on the Random Forest confidence score rather than producing their confidence as a function of the data itself, have no way of adapting sufficiently to the extreme changes in the data and therefore have no way of anticipating such a drastic decrease in model performance.

MACEst on the other hand continues to follow the trend in accuracy and generally indicates a more appropriate level of confidence relative to the performance of the model. 
In particular, when the noise becomes large enough, MACEst is able to report that the model is returning predictions which are no better than random guesses. 

This is incredibly important when using models in real-world applications as often beyond the training data, it is hard to have immediate feedback regarding the outcome of any given prediction. 
This means that it is generally non-trivial to monitor the model performance live, so the model confidence may be the only metric available in real-time. 
If one were to use that metric in this example, beyond the noise level with a standard deviation of 2, then unless MACEst was used, there would be no indication that the data had changed and that the model is getting less capable at making good predictions.

\subsection{Is MACEst trustworthy?}
 We now turn to our final critique of traditional calibration models.
 We have shown that often the correlation between confidence and trust scores can be broken when data is sufficiently different to the training data. 
 This indicates that traditional calibration methods are not accounting for epistemic uncertainty and are therefore not necessarily trustworthy. 
 We now repeat the experiment in section \ref{sec: not trustworthy} with MACEst and compare the behaviour between MACEst and traditional calibration methods. 

\begin{figure*} [t]
        \centering
        \includegraphics[width= 0.95\textwidth]{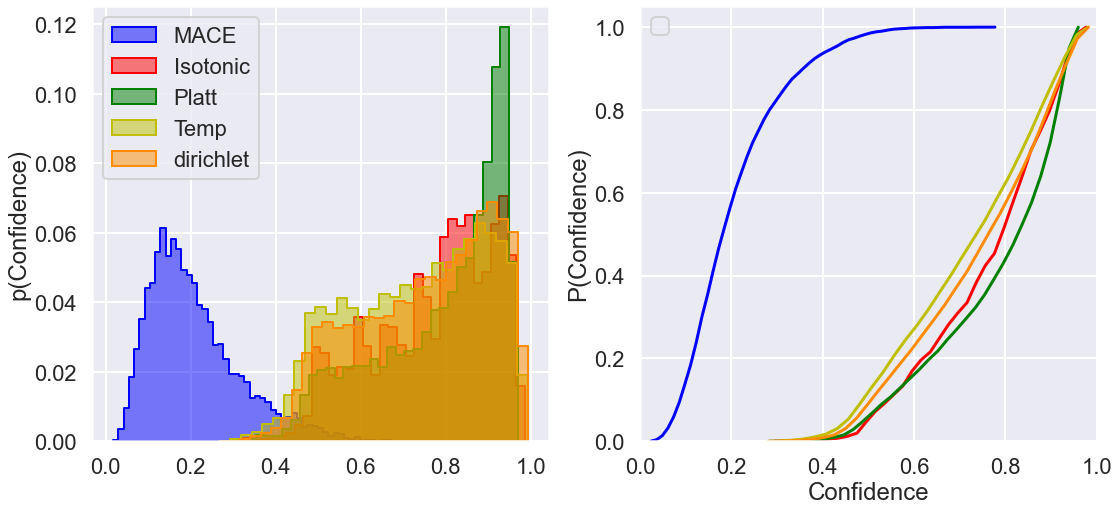}
        \caption{Comparison of the distribution of confidence predictions on random noise.
        We show the probability distribution function (left) and the smoothed cumulative density function (right). 
        MACEst is shown ton be significantly less confident than the other methods with many predictions returning $\sim 0.1 $ corresponding to no clear information and effectively zero high confident predictions.}
        \label{fig:mace_noise_dist}
\end{figure*}

As shown in fig \ref{fig:mace_noise_dist} when MACEst is asked to make predictions on random noise, the distribution clearly changes relative to the calibration methods. 
We quantify this difference with the Kolmogorov–Smirnov (KS) statistic and find that MACEst vs. Isotonic, Platt, Temperature, Dirichlet scores are $ 0.95 \pm 0.02$. 
This significant change is because MACEst's distribution tends towards very low confidence predictions. 
We find that often MACEst returns probabilities of around $0.1$ indicating that the model is not able to make a clear prediction.

\begin{figure*} [t]
        \centering
        \includegraphics[width= 0.7\textwidth]{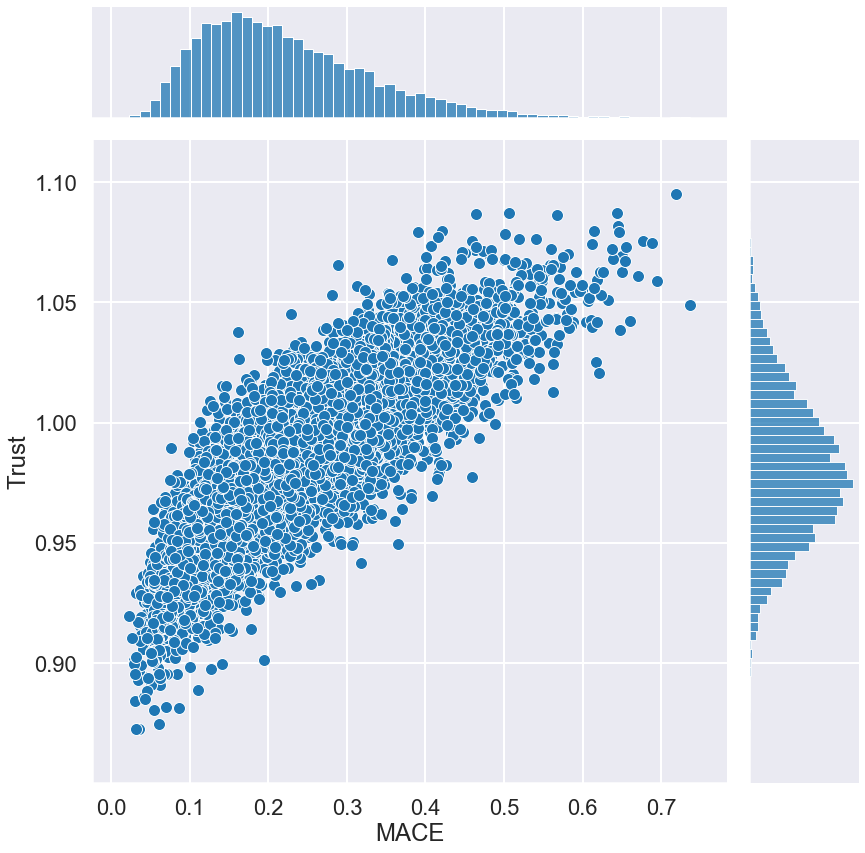}
        \caption{The joint distribution between confidence estimates and trust scores for MACEst. 
        We see that, unlike other methods, there is a very clear correlation between the two indicating that MACEst remains trustworthy under large epistemic uncertainty.}
        \label{fig:mace_trust_score}
\end{figure*}

We now compare MACEst and trust scores.
As expected when we are making predictions on data that is similar to the training data confidence and trust scores are generally both very high (Spearman rank correlation = $0.964$). 
In Figure \ref{fig:mace_trust_score} of the joint distribution between MACEst confidence estimates and trust scores when making predictions on random noise we see that MACEst and trust scores remain highly correlated (correlation = $0.841$).
This indicates that MACEst is correctly incorporating epistemic uncertainty and therefore remains trustworthy despite the data being entirely different to the data which the point prediction model was trained on. 
See table \ref{tab:trust_conf_corr} for a full comparison between each calibration method.

\begin{table}[]
\centering
\begin{tabular}{|l|l|l|}
\hline
                     & \textbf{In sample test data} & \textbf{Out of sample Noise} \\ \hline
\textbf{MACEst}        & 0.964                        & 0.841                        \\ \hline
\textbf{Isotonic}     & 0.797                        & 0.056                        \\ \hline
\textbf{Platt}       & 0.803                        & 0.085                        \\ \hline
\textbf{Temperature} & 0.814                        & 0.108                        \\ \hline
\textbf{Dirichlet}   & 0.773                        & 0.089                        \\ \hline
\end{tabular}
\caption{Table comparing the Spearman rank correlation between trust scores and confidence estimates for MNIST data. 
The first column shows the correlation when estimating both quantities on an unseen test set. 
The second column shows the correlation when calculating each quantity on random noise}
\label{tab:trust_conf_corr}
\end{table}

\section{Discussion}
The results from our experiments suggest that despite MACEst being an incredibly simple model, a linear sum of two derived quantities (there are likely more sophisticated versions based on this philosophy that may be better), it is comparable with the state of the art across a variety of metrics whilst also not suffering from the flaws highlighted in the methods based upon scaling point prediction scores. 
We will first address some caveats and limitations of this method before pointing to several potential important applications of MACEst and looking forward to future work in this area.

\subsection{Potential Limitations}
Firstly, MACEst relies on the assumption that it is possible to define a notion of similarity between data points, however the distance between raw features is not necessarily a good measure of similarity and it is likely that more care will be needed in the data cleaning and feature engineering steps - this is evidenced in its relatively poor performance in NLP use cases.
This can be equivalently expressed as the need to embed the data into a sensible co-ordinate space so that a similarity measure is appropriate.
Embedding into a specific distance metric area is a very active area of research (see \cite{kaya2019deep} for a recent survey).

The computational costs associated with MACEst are predominantly in the neighbour search, exact nearest neighbour search is $ O(n^2)$ that might be expensive for any reasonably large dataset. 
Despite our use of efficient approximate nearest neighbour search the cost of building an additional nearest neighbour model may still be prohibitive for some applications.

In terms of data efficiency, MACEst does not have many parameters to train so does not require a large amount of data to tune them; however sufficient data is needed to train the point prediction model, as well as large graph data, and then the calibration and test sets must be sufficiently large to be useful, i.e. to induce the graph space and generate representative accuracy estimations.
As a very rough rule of thumb we found that in order to have sufficient data to train both MACEst and the point prediction model well, we required several thousand or more data points for the above classification tasks.

\subsection{Practical Applications }
As well as the intrinsic utility of getting good confidence estimates we believe that the intuitive nature of MACEst means that it has other important uses in the data science community. 
Firstly, the confidence estimates have a clear interpretation. 
If you have low confidence you either don't have enough relevant data or the model predictions in this region are noisy or error-prone. 
This can easily be translated into useful insights for human in the loop pipelines, i.e. seeking to understand why the model struggles to predict well and/or collect more data in certain regions. 

Anomaly detection is also trivial when using MACEst, any anomalous data point will by definition be very different to the training data. 
This will mean that MACEst predicts a large uncertainty, and in particular a large epistemic uncertainty. 
This could easily be converted into an automated test, for which points with epistemic uncertainty larger than some threshold are automatically flagged.

Another issue that practitioners face is how to effectively monitor a model and when a model will need to be re-trained, this is related to the issue of data-drift over time. 
This is the notion that over time data will change and models will become less effective. 
Models must then be re-trained on new data, however re-training models can be costly and when to re-train is a somewhat arbitrary choice because there are generally not good metrics for \emph{goodness of data} making this choice particularly difficult (\cite{sambasivan2021everyone}). 
MACEst again provides a very simple solution to this problem, as shown in section \ref{sec: epistemic}, as the data becomes more different to the initial training data the confidence estimates will on average decrease.
This means there is a very clear metric to track when considering how well a model is performing. 
This is a better metric than predictive accuracy because the reason for confidence decline will be more clearly related to the data-drift. 
This could lead to simple automation rules such as re-train if average confidence falls below x\%. 

\subsection{Future work}
There are some problems where knowledge of a single feature of many is enough to be very confident about a prediction, this concept of feature importance and feature interaction is not explicitly accounted for in MACEst currently. 
Accounting for feature importance is a common problem in machine learning applications and is an active area of research (\cite{fisher2019all, huang2019fibinet}). 
Pre-processing and feature engineering can somewhat mitigate these problems, however in future work we plan to explicitly incorporate feature interactions and feature importance into any similarity metric.

This study has not explicitly looked at applications in deep learning, this is because MACEst is a general method which is supposed to be model agnostic. 
Deep learning is effective when tackling problems where the raw features themselves are not necessarily good predictors, and so the network is able to induce new features. 
This of course means the notion of a similarity metric between two data points is more challenging.

In future work we plan to explore the applications of MACEst to deep learning, in particular we plan to compare methods which will induce a similarity metric such as Auto-encoders, embeddings, etc. 
In principle as long as there is, or one can induce, a good measure of similarity then MACEst should be effective. 
Combining metric learning and similarity networks \cite{kaya2019deep} with MACEst would make this method applicable to a wider range of problems.

The applications to deep learning are important because many of the problems highlighted above, in particular decision boundaries and extrapolation are even bigger issues for deep learning applications. 
The most extreme version of this being adversarial attacks (\cite{szegedy2013intriguing}). 
In principle MACEst will offer a very robust counter-measure to adversarial attacks for the same reasons that MACEst is robust to high confidence predictions on pure noise.

This paper has presented MACEst as working in the context of classification algorithms, however the principles are also applicable to regression and a very similar study into using MACEst for prediction intervals is in preparation.

\section{Conclusion}
Modern Machine learning algorithms are incredibly effective at classification tasks: in this paper we have argued and evidenced that despite this such algorithms are not a good starting point for computing confidence estimates. 
Our position is that any confidence estimates based upon this paradigm inevitably fail to account for epistemic uncertainty that is crucial to producing reliable confidence estimates. 
This often leads to \emph{over-confident} predictions on data points that are dissimilar to the data used to train the point prediction model. 
To counteract this, in this work we presented a simple alternative, MACEst, that is based upon a fundamentally different paradigm.
Using a set of nearest neighbours, MACEst estimates uncertainty locally and explicitly accounts for both epistemic and aleatoric uncertainty. 
MACEst can be applied as an ad-hoc step in any machine learning pipeline to provide accurate and robust confidence estimates. 
We have shown that in many situations the confidence estimates produced by MACEst will be more reliable and are therefore more suitable than other methods for most practical applications.

\acks{We are grateful to Dan Golding, Hugo Monteiro, Oleg Shevelyov for code review and testing and Steve Fairhurst for useful comments on the manuscript. RG was supported by Science and Technology Facilities Council (STFC) grant ST/L000962/1.}

\newpage
\bibliography{mace_refs}

\begin{thebibliography}{39}
\providecommand{\natexlab}[1]{#1}
\providecommand{\url}[1]{\texttt{#1}}
\expandafter\ifx\csname urlstyle\endcsname\relax
  \providecommand{\doi}[1]{doi: #1}\else
  \providecommand{\doi}{doi: \begingroup \urlstyle{rm}\Url}\fi

\bibitem[Aum{\"u}ller et~al.(2017)Aum{\"u}ller, Bernhardsson, and
  Faithfull]{aumuller2017ann}
Martin Aum{\"u}ller, Erik Bernhardsson, and Alexander Faithfull.
\newblock Ann-benchmarks: A benchmarking tool for approximate nearest neighbor
  algorithms.
\newblock In \emph{International Conference on Similarity Search and
  Applications}, pages 34--49. Springer, 2017.

\bibitem[Bishop(2006)]{BishopPatternrecognitionmachine2006}
Christopher~M. Bishop.
\newblock \emph{Pattern Recognition and Machine Learning}.
\newblock Information Science and Statistics. {Springer}, {New York}, 2006.
\newblock ISBN 978-0-387-31073-2.

\bibitem[Breiman(2001)]{breiman2001random}
Leo Breiman.
\newblock Random forests.
\newblock \emph{Machine learning}, 45\penalty0 (1):\penalty0 5--32, 2001.

\bibitem[Congdon(2014)]{congdon2014applied}
Peter Congdon.
\newblock \emph{Applied bayesian modelling}, volume 595.
\newblock John Wiley \& Sons, 2014.

\bibitem[Cortes and Vapnik(1995)]{cortes1995support}
Corinna Cortes and Vladimir Vapnik.
\newblock Support-vector networks.
\newblock \emph{Machine learning}, 20\penalty0 (3):\penalty0 273--297, 1995.

\bibitem[Devlin et~al.(2018)Devlin, Chang, Lee, and Toutanova]{devlin2018bert}
Jacob Devlin, Ming-Wei Chang, Kenton Lee, and Kristina Toutanova.
\newblock Bert: Pre-training of deep bidirectional transformers for language
  understanding.
\newblock \emph{arXiv preprint arXiv:1810.04805}, 2018.

\bibitem[Dua and Graff(2017)]{Dua:2019}
Dheeru Dua and Casey Graff.
\newblock {UCI} machine learning repository, 2017.
\newblock URL \url{http://archive.ics.uci.edu/ml}.

\bibitem[Fisher et~al.(2019)Fisher, Rudin, and Dominici]{fisher2019all}
Aaron Fisher, Cynthia Rudin, and Francesca Dominici.
\newblock All models are wrong, but many are useful: Learning a variable's
  importance by studying an entire class of prediction models simultaneously.
\newblock \emph{Journal of Machine Learning Research}, 20\penalty0
  (177):\penalty0 1--81, 2019.

\bibitem[Friedman et~al.(2001)Friedman, Hastie, and
  Tibshirani]{friedman2001elements}
Jerome Friedman, Trevor Hastie, and Robert Tibshirani.
\newblock \emph{The elements of statistical learning}, volume~1.
\newblock Springer series in statistics New York, 2001.

\bibitem[Gal and Ghahramani(2015)]{gal_dropout_2015}
Yarin Gal and Zoubin Ghahramani.
\newblock Dropout as a {{Bayesian Approximation}}: {{Representing Model
  Uncertainty}} in {{Deep Learning}}.
\newblock \emph{arXiv:1506.02142 [cs, stat]}, June 2015.

\bibitem[Gneiting and Raftery(2007)]{gneiting2007strictly}
Tilmann Gneiting and Adrian~E Raftery.
\newblock Strictly proper scoring rules, prediction, and estimation.
\newblock \emph{Journal of the American statistical Association}, 102\penalty0
  (477):\penalty0 359--378, 2007.

\bibitem[Goldberger et~al.(2004)Goldberger, Hinton, Roweis, and
  Salakhutdinov]{goldberger2004neighbourhood}
Jacob Goldberger, Geoffrey~E Hinton, Sam Roweis, and Russ~R Salakhutdinov.
\newblock Neighbourhood components analysis.
\newblock \emph{Advances in neural information processing systems},
  17:\penalty0 513--520, 2004.

\bibitem[Guo et~al.(2017)Guo, Pleiss, Sun, and
  Weinberger]{GuoCalibrationModernNeural2017}
Chuan Guo, Geoff Pleiss, Yu~Sun, and Kilian~Q. Weinberger.
\newblock On {{Calibration}} of {{Modern Neural Networks}}.
\newblock \emph{arXiv:1706.04599 [cs]}, August 2017.

\bibitem[Guo et~al.(2020)Guo, Sun, Lindgren, Geng, Simcha, Chern, and
  Kumar]{avq_2020}
Ruiqi Guo, Philip Sun, Erik Lindgren, Quan Geng, David Simcha, Felix Chern, and
  Sanjiv Kumar.
\newblock Accelerating large-scale inference with anisotropic vector
  quantization.
\newblock In \emph{International Conference on Machine Learning}, 2020.
\newblock URL \url{https://arxiv.org/abs/1908.10396}.

\bibitem[Hinton and Salakhutdinov(2006)]{hinton2006reducing}
Geoffrey~E Hinton and Ruslan~R Salakhutdinov.
\newblock Reducing the dimensionality of data with neural networks.
\newblock \emph{science}, 313\penalty0 (5786):\penalty0 504--507, 2006.

\bibitem[Huang et~al.(2019)Huang, Zhang, and Zhang]{huang2019fibinet}
Tongwen Huang, Zhiqi Zhang, and Junlin Zhang.
\newblock Fibinet: combining feature importance and bilinear feature
  interaction for click-through rate prediction.
\newblock In \emph{Proceedings of the 13th ACM Conference on Recommender
  Systems}, pages 169--177, 2019.

\bibitem[Jiang et~al.(2018)Jiang, Kim, Guan, and Gupta]{JiangTrustNotTrust2018}
Heinrich Jiang, Been Kim, Melody~Y. Guan, and Maya Gupta.
\newblock To {{Trust Or Not To Trust A Classifier}}.
\newblock \emph{arXiv:1805.11783 [cs, stat]}, May 2018.

\bibitem[Johnson et~al.(2017)Johnson, Douze, and J{\'e}gou]{JDH17}
Jeff Johnson, Matthijs Douze, and Herv{\'e} J{\'e}gou.
\newblock Billion-scale similarity search with gpus.
\newblock \emph{arXiv preprint arXiv:1702.08734}, 2017.

\bibitem[Jolliffe(2003)]{jolliffe2003principal}
IT~Jolliffe.
\newblock Principal component analysis.
\newblock \emph{Technometrics}, 45\penalty0 (3):\penalty0 276, 2003.

\bibitem[Kaya and Bilge(2019)]{kaya2019deep}
Mahmut Kaya and Hasan~{\c{S}}akir Bilge.
\newblock Deep metric learning: A survey.
\newblock \emph{Symmetry}, 11\penalty0 (9):\penalty0 1066, 2019.

\bibitem[Kendall and Gal(2017)]{kendall2017uncertainties}
Alex Kendall and Yarin Gal.
\newblock What uncertainties do we need in bayesian deep learning for computer
  vision?
\newblock In \emph{Advances in neural information processing systems}, pages
  5574--5584, 2017.

\bibitem[Kull et~al.(2019)Kull, Perello~Nieto, K{\"a}ngsepp, Silva~Filho, Song,
  and Flach]{kull_beyond_2019}
Meelis Kull, Miquel Perello~Nieto, Markus K{\"a}ngsepp, Telmo Silva~Filho, Hao
  Song, and Peter Flach.
\newblock Beyond temperature scaling: {{Obtaining}} well-calibrated multi-class
  probabilities with {{Dirichlet}} calibration.
\newblock In H.~Wallach, H.~Larochelle, A.~Beygelzimer,
  F.~d\textbackslash{}textquotesingle {Alch{\'e}-Buc}, E.~Fox, and R.~Garnett,
  editors, \emph{Advances in {{Neural Information Processing Systems}} 32},
  pages 12295--12305. {Curran Associates, Inc.}, 2019.

\bibitem[Lakshminarayanan et~al.(2017)Lakshminarayanan, Pritzel, and
  Blundell]{lakshminarayanan2017simple}
Balaji Lakshminarayanan, Alexander Pritzel, and Charles Blundell.
\newblock Simple and scalable predictive uncertainty estimation using deep
  ensembles.
\newblock In \emph{Advances in neural information processing systems}, pages
  6402--6413, 2017.

\bibitem[LeCun et~al.(2010)LeCun, Cortes, and Burges]{lecun2010mnist}
Yann LeCun, Corinna Cortes, and CJ~Burges.
\newblock Mnist handwritten digit database.
\newblock \emph{ATT Labs [Online]. Available:
  http://yann.lecun.com/exdb/mnist}, 2, 2010.

\bibitem[Loh(2014)]{loh2014fifty}
Wei-Yin Loh.
\newblock Fifty years of classification and regression trees.
\newblock \emph{International Statistical Review}, 82\penalty0 (3):\penalty0
  329--348, 2014.

\bibitem[Malkov and Yashunin(2018)]{malkov2018efficient}
Yury~A Malkov and Dmitry~A Yashunin.
\newblock Efficient and robust approximate nearest neighbor search using
  hierarchical navigable small world graphs.
\newblock \emph{IEEE transactions on pattern analysis and machine
  intelligence}, 2018.

\bibitem[Mikolov et~al.(2013)Mikolov, Sutskever, Chen, Corrado, and
  Dean]{mikolov2013distributed}
Tomas Mikolov, Ilya Sutskever, Kai Chen, Greg~S Corrado, and Jeff Dean.
\newblock Distributed representations of words and phrases and their
  compositionality.
\newblock In \emph{Advances in neural information processing systems}, pages
  3111--3119, 2013.

\bibitem[Murphy and Winkler(1977)]{murphy1977reliability}
Allan~H Murphy and Robert~L Winkler.
\newblock Reliability of subjective probability forecasts of precipitation and
  temperature.
\newblock \emph{Journal of the Royal Statistical Society: Series C (Applied
  Statistics)}, 26\penalty0 (1):\penalty0 41--47, 1977.

\bibitem[Naeini et~al.(2015)Naeini, Cooper, and
  Hauskrecht]{naeini2015obtaining}
Mahdi~Pakdaman Naeini, Gregory Cooper, and Milos Hauskrecht.
\newblock Obtaining well calibrated probabilities using bayesian binning.
\newblock In \emph{Twenty-Ninth AAAI Conference on Artificial Intelligence},
  2015.

\bibitem[Neal(2012)]{neal2012bayesian}
Radford~M Neal.
\newblock \emph{Bayesian learning for neural networks}, volume 118.
\newblock Springer Science \& Business Media, 2012.

\bibitem[{Niculescu-Mizil} and
  Caruana(2005)]{Niculescu-MizilPredictinggoodprobabilities2005}
Alexandru {Niculescu-Mizil} and Rich Caruana.
\newblock Predicting good probabilities with supervised learning.
\newblock In \emph{Proceedings of the 22nd International Conference on
  {{Machine}} Learning - {{ICML}} '05}, pages 625--632, {Bonn, Germany}, 2005.
  {ACM Press}.
\newblock ISBN 978-1-59593-180-1.
\newblock \doi{10.1145/1102351.1102430}.

\bibitem[Nixon et~al.(2019)Nixon, Dusenberry, Zhang, Jerfel, and
  Tran]{nixon2019measuring}
Jeremy Nixon, Michael~W Dusenberry, Linchuan Zhang, Ghassen Jerfel, and Dustin
  Tran.
\newblock Measuring calibration in deep learning.
\newblock In \emph{CVPR Workshops}, pages 38--41, 2019.

\bibitem[Platt et~al.(1999)]{platt1999probabilistic}
John Platt et~al.
\newblock Probabilistic outputs for support vector machines and comparisons to
  regularized likelihood methods.
\newblock \emph{Advances in large margin classifiers}, 10\penalty0
  (3):\penalty0 61--74, 1999.

\bibitem[Rasmussen(2003)]{rasmussen2003gaussian}
Carl~Edward Rasmussen.
\newblock Gaussian processes in machine learning.
\newblock In \emph{Summer School on Machine Learning}, pages 63--71. Springer,
  2003.

\bibitem[Sambasivan et~al.(2021)Sambasivan, Kapania, Highfill, Akrong,
  Paritosh, and Aroyo]{sambasivan2021everyone}
Nithya Sambasivan, Shivani Kapania, Hannah Highfill, Diana Akrong,
  Praveen~Kumar Paritosh, and Lora~Mois Aroyo.
\newblock " everyone wants to do the model work, not the data work": Data
  cascades in high-stakes ai.
\newblock 2021.

\bibitem[Szegedy et~al.(2013)Szegedy, Zaremba, Sutskever, Bruna, Erhan,
  Goodfellow, and Fergus]{szegedy2013intriguing}
Christian Szegedy, Wojciech Zaremba, Ilya Sutskever, Joan Bruna, Dumitru Erhan,
  Ian Goodfellow, and Rob Fergus.
\newblock Intriguing properties of neural networks.
\newblock \emph{arXiv preprint arXiv:1312.6199}, 2013.

\bibitem[Tipping(2003)]{tipping2003bayesian}
Michael~E Tipping.
\newblock Bayesian inference: An introduction to principles and practice in
  machine learning.
\newblock In \emph{Summer School on Machine Learning}, pages 41--62. Springer,
  2003.

\bibitem[Williams and Rasmussen(2006)]{williams2006gaussian}
Christopher~KI Williams and Carl~Edward Rasmussen.
\newblock \emph{Gaussian processes for machine learning}, volume~2.
\newblock MIT press Cambridge, MA, 2006.

\bibitem[Zadrozny and Elkan(2002)]{ZadroznyTransformingClassifierScores2002}
Bianca Zadrozny and Charles Elkan.
\newblock \emph{Transforming {{Classifier Scores}} into {{Accurate Multiclass
  Probability Estimates}}}.
\newblock 2002.

\end{thebibliography}

\end{document}